\documentclass[conference]{IEEEtran}
\IEEEoverridecommandlockouts
\usepackage{cite}
\usepackage{amsmath,amssymb,amsfonts}
\usepackage{algorithmic}
\usepackage{graphicx}
\usepackage{textcomp}
\usepackage{xcolor}
\def\BibTeX{{\rm B\kern-.05em{\sc i\kern-.025em b}\kern-.08em
    T\kern-.1667em\lower.7ex\hbox{E}\kern-.125emX}}
\begin{document}

\title{Detection and Classification of Industrial Signal Lights for Factory Floors
\thanks{ This work has been carried out by Felix Nilsson in the context of his Bachelor Thesis in Computer Science and Engineering at Halmstad University. Supported by HMS Industrial Networks AB.}
}

\author{\IEEEauthorblockN{Felix Nilsson}
\IEEEauthorblockA{\textit{HMS Industrial Networks AB} \\
Halmstad, Sweden  \\
fenil@hms.se}
\and
\IEEEauthorblockN{Jens Jakobsen}
\IEEEauthorblockA{\textit{HMS Industrial Networks AB} \\
Halmstad, Sweden  \\
jeja@hms.se}
\and
\IEEEauthorblockN{Fernando Alonso-Fernandez}
\IEEEauthorblockA{\textit{School of Information Technology (ITE)} \\
\textit{Halmstad University}, Sweden  \\
feralo@hh.se}
}

\maketitle

\begin{abstract}
Industrial manufacturing has developed during the last decades from a labor-intensive manual control of machines to a fully-connected automated process. The next big leap is known as industry 4.0, or smart manufacturing. With industry 4.0 comes increased integration between IT systems and the factory floor from the customer order system to final delivery of the product. One benefit of this integration is mass production of individually customized products. However, this has proven challenging to implement into existing factories, considering that their lifetime can be up to 30 years. The single most important parameter to measure in a factory is the operating hours of each machine. Operating hours can be affected by machine maintenance as well as re-configuration for different products. For older machines without connectivity, the operating state is typically indicated by signal lights of green, yellow and red colours. 
Accordingly, the goal is to develop a solution which can measure the operational state using the input from a video camera capturing a factory floor.
Using methods commonly employed for traffic light recognition in autonomous cars, a system with an accuracy of over 99\% in the specified conditions is presented. It is believed that if more diverse video data becomes available, a system with high reliability that generalizes well could be developed using a similar methodology.
\end{abstract}


\section{Introduction}

\begin{figure*}
\centering
\includegraphics[width=0.85\textwidth]{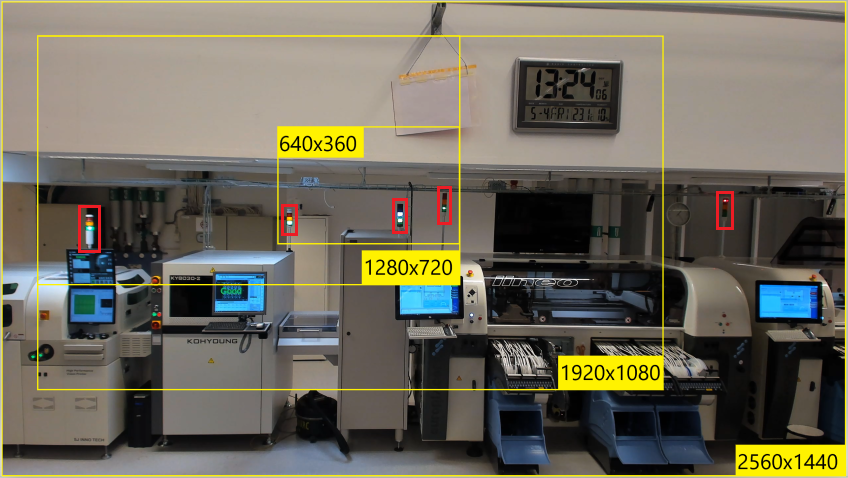}
\includegraphics[width=0.85\textwidth]{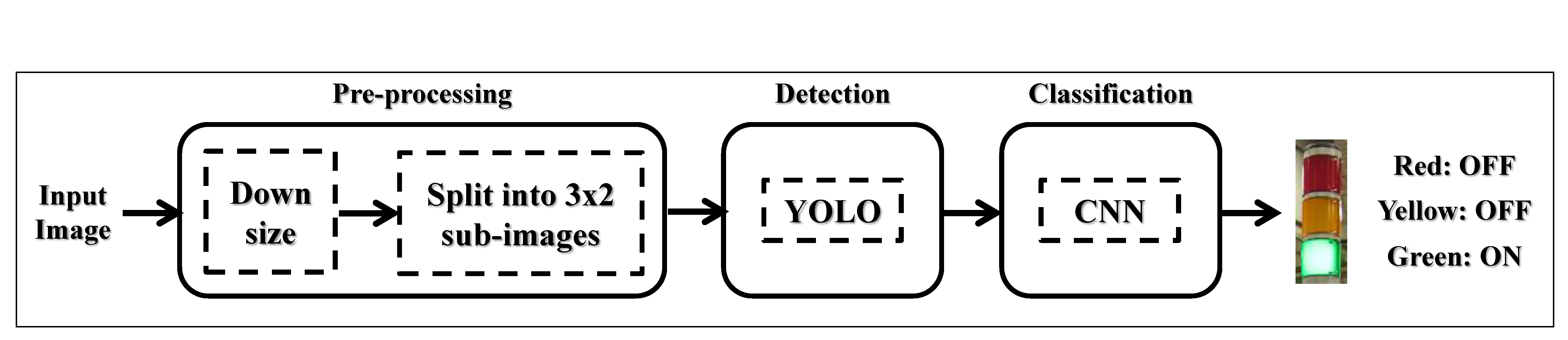}
\caption{Top: Factory floor of our acquisition setup. Bottom: Overview of our system.}
\label{fig:example-stack-light-and-system}
\end{figure*}

The current trend in industrial manufacturing is commonly referred to as Industry 4.0, or smart manufacturing \cite{[Hermann16_industry40]}.
In Industry 4.0, the factory floor is increasingly integrated with IT systems from the customer order to the delivery of the final product.
It involves interconnecting the different stages of the manufacturing process with IT systems, entailing the processing and storage of large amounts of data.
%
Implementing Industry 4.0 into existing factories that lack connectivity with IT systems is a great challenge, especially when considering that the lifetime of a factory can be up to 30 years.
An important parameter in industrial manufacturing is the operating hours of each machine.
On machines, the operating state is typically indicated by a stack light or industrial signal light (Figure~\ref{fig:example-stack-light-and-system}).
Green indicates running, red indicates an error, and yellow/amber indicates idle state.
Automated monitoring of the machine state is a vital tool for optimization methods such as lean manufacturing and Six Sigma.
In addition, an alarm message can be generated in case of machine malfunction.

%
%
%

Existing solutions to track the operational state rely on cables being drawn to the machines. This is costly and cumbersome, and might not even be an option in older machines that lack the interfaces needed. 
Therefore, the goal of this work is to develop a system which makes use of a video feed from the factory floor. The system shall first locate and then classify the signal lights belonging to each machine.
This way, the information about the operational state could be collected automatically and transparently without impact to the rest of the factory infrastructure.
One detection and one classification network based on YOLOv2 \cite{[Redmon17YOLOv2]} and AlexNet \cite{[Krizhevsky12]} respectively have been trained and evaluated for this purpose.
Although popular object detection architectures such as YOLO or R-CNN could be also trained for classification \cite{[Redmon16YOLO],[Redmon17YOLOv2],[Girshick14RCNN]}, it would require very powerful hardware.
The proposed solution allows the use of a simpler classification architecture, and also enables to evaluate detection and classification performance separately.
A dataset with 125613 frames and 628095 sub-images of stack lights have been captured as well, which are used for training and evaluation.

\subsection{Related Works}

Despite being a relevant issue, the literature on industrial signal light detection or classification is scarce. A search on IEEE Xplore, ScienceDirect and SpringerLink (the major sources of publications in engineering) only produces one result \cite{[Matare17industrialligths]}.
Here, detection is done by colour information to determine regions of interest. Regions are then combined according to some spatial constraints to eliminate false positives.
The process is aided by a a 2D laser scanner that provides depth information, since the solution is evaluated in the context of moving machines and cameras.
The tasks at hand share many similarities with recognition of traffic lights in autonomous cars, which is a very active field.
For this reason, we have looked into the literature of this field as well.
Traffic light recognition can be done in a variety of ways, including broadcasting traffic signal states over radio, but this would require major investments in infrastructure.
For this reason, the majority of works rely solely on video cameras. 
Nowadays, the majority of researchers tend to use deep learning and neural networks. Some older articles from around ten plus years ago propose different solutions.

In the work \cite{[Charette09trafficlights]}, the authors suggest a method with three steps.
The first is Spot Light Detection, referred to as SLD, to detect candidate regions.
An algorithm identifies the bright areas and then, by shape filtering, only keeps those that have the shape and proportions of a lens.
Secondly, the algorithm tries to identify the housing and pole (which have the shape of a rectangle),
and the bright spots themselves inside the housing.
The shapes are evaluated in terms of size, proportions and relative positions.
Lastly, the state of the light is identified from the position of the bright spot inside the housing (red, yellow, green, from top to bottom).
The authors of \cite{[Kim07trafficlights]} take a different approach. Here,
the red, yellow and green planes are put through a thresholding algorithm, which allows to derive the colour of the light.
However, this solution has a lower alleged accuracy in comparison to the method previously mentioned, 90\% versus 97\%.

As mentioned, deep learning has gained popularity in object recognition.
Initially, it was applied only in the classification of traffic lights, with candidates first detected using
classical methods like colour segmentation, aspect ratio analysis, and other parameters \cite{[Saini17trafficlights],[Lee17trafficlights]}.
%
In recent works, deep learning is employed both for detection and classification.
They are based on popular object detection architectures such as YOLO \cite{[Redmon16YOLO]}
or R-CNN \cite{[Girshick14RCNN]} and its derivatives (Fast R-CNN, Faster R-CNN, and R-FCN).
In the work \cite{[Kim18trafficlights]},
three types of ensemble network models based on Faster R-CNN and R-FCN are tested together with six different colour spaces (e.g. RGB, YCbCr).
Candidate regions are selected using Inception-Resnet-v2 or Resnet-101 as feature extractors, and these are then fed
to Faster R-CNN or R-FCN for classification.
%
%
%
YOLO is used for traffic light recognition in other works, both on its own \cite{[Jensen17trafficlight]},
and in conjunction with a smaller classification network to improve accuracy \cite{[Behrendt17trafficlights]}.
YOLO has the advantage of being much faster than R-CNNs, but still having similar accuracy.
%

\begin{figure}
\centering
\includegraphics[width=0.45\textwidth]{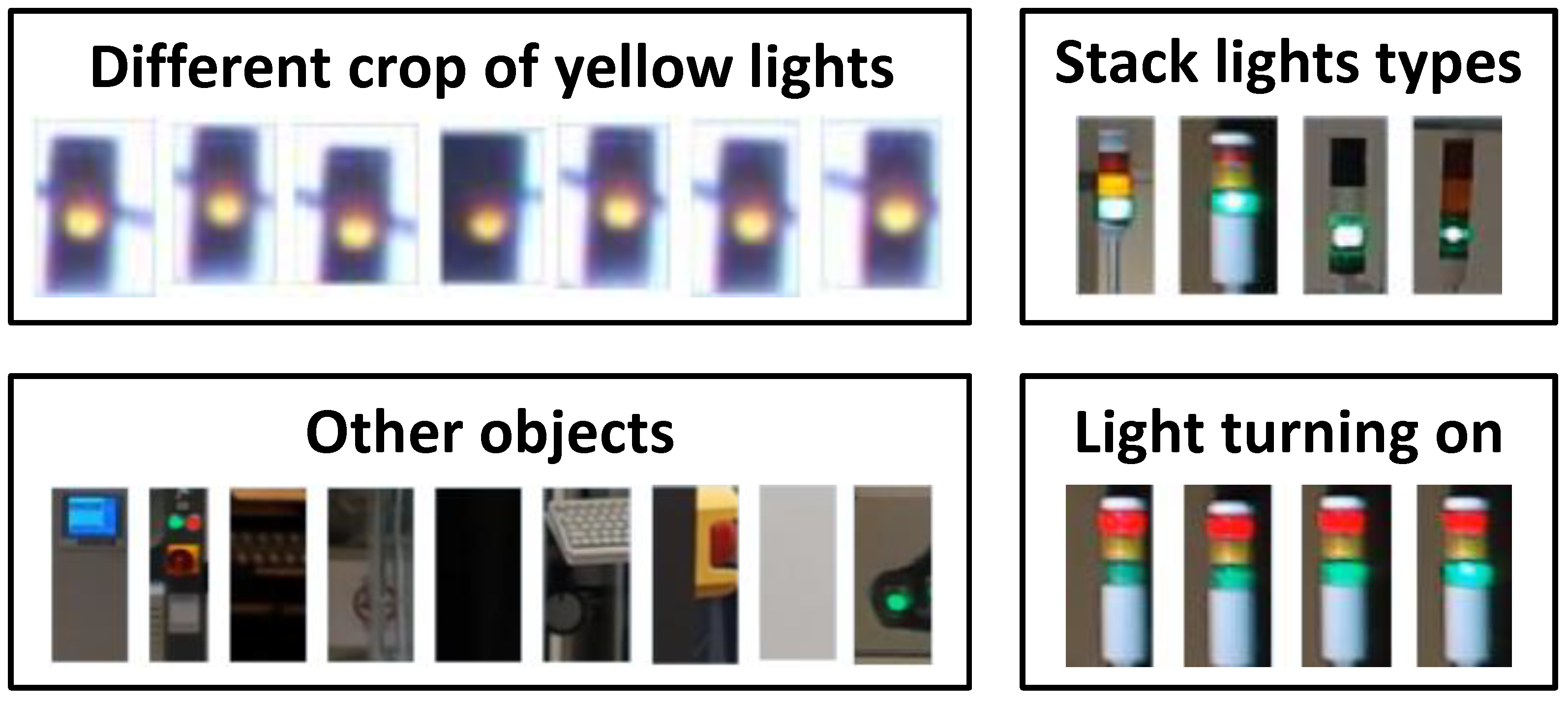}
\caption{Top/Left: different crop of yellow lights (TLD datasets). Top/Right: types of lights in our stack lights dataset.
        Bottom/Left: category with ``other'' objects. Bottom/Right: example of a stack light turning on to green.}
\label{fig:datasets}
\end{figure}

\section{Methodology}

\subsection{System Overview}
\label{sec:system}

The proposed system has three steps, as shown in Figure~\ref{fig:example-stack-light-and-system}, bottom: pre-processing, detection, and classification.
As detector, we employ the YOLOv2 network \cite{[Redmon17YOLOv2]}. It has an input size of 416$\times$416.
Although a higher resolution is possible, both training and running would be slow and require a large amount of GPU power and memory.
Since the images of our dataset have been captured in QHD resolution (2560$\times$1440),
we carry out some pre-processing to avoid that objects of interest become too small when downsizing to 416$\times$416.
%
%
%
%
Input images are downsized to 1248$\times$832 pixels. This resolution is detailed enough for our objects of interest, and divisible by 416 on both axes.
Images are then split into six sub-images of 416$\times$416 pixels, which are sent through the detection network separately.
YOLO outputs bounding boxes, some of which may overlap each other due to multiple detections caused by the image being split.
To handle, this non-maximal suppression is performed.
If there are several bounding boxes with an overlap greater than 50\%,
only the bounding box with the highest prediction score is kept.
%
%
The remaining bounding boxes are then passed on for classification to a CNN.
For this, a small network is used (a modified version of AlexNet \cite{[Krizhevsky12]}, which is 8 layers deep),
with an input image size of 227$\times$227.
The bounding boxes given by the detector are resized to this dimension,
stretching them in the smallest dimension (width) so that they become square.
The objects to be classified have low complexity (colours), and our system will operate indoors with fixed cameras and machines.
Therefore, a more complex network is not necessary.
If our system would need to handle with outdoor conditions or motion, a deeper architecture might be used instead.
However, a shallower network is sufficient for our system, which results in faster classification speeds as well.

Detection and classification are deliberately kept separate. Detection is more costly in processing power
with the networks employed, and since machines and cameras are fixed in our scenario,
it is not necessary in every frame.
If this would not be the case, the interval at which detection is performed could be changed.
In our case it is invoked every 0.2 seconds to detect potential occlusions, such as people passing by.
%
Keeping both steps separate also allows to train and evaluate them individually, with has advantages if adaptation to more adverse environments is needed.
In such case, it is expected that just improving the detector to deal with more difficult imaging might be sufficient.
%
Given the availability of data in the field of autonomous driving,
this has also allowed to pre-train the classification network with a richer dataset of traffic lights from traffic scenes.

\subsection{Datasets}

Data from three different sources have been used in this paper.

\subsubsection{Traffic Light Datasets (TLD)}
%
cropped from traffic scenes, and initially employed to train the classification CNN.
For this, the LISA Traffic Light \cite{[Philipsen15_LISATLdataset]} and BSTLD Bosch Small Traffic Lights Dataset \cite{[Behrendt17trafficlights]} was used.
The lights are cropped with a margin of some pixels to enable context awareness.
Table~\ref{tab:TLDdbs} shows the distribution of data.
Since yellow is severely under-represented,
yellow lights were cropped several times
with a random offset of $\pm$5 pixels in vertical and horizontal
directions.
Figure~\ref{fig:datasets} (Top/Left) shows an example.
The resulting dataset consists of 61712 images.

\begin{table}[htb]
  \begin{center}
    \caption{Distribution of traffic lights between categories (TLD datasets).}
    \label{tab:TLDdbs}
    \begin{tabular}{|c||c|c|c||c|c|}
  \hline
  Colour & BSTLD & LISA & (\%) & Adjusted & (\%) \\ \hline
  Green & 7566 & 13830 & 50.1\% & 21396 & 34.7\% \\
  Yellow & 154 & 755 & 2.1\% & 19889 & 32.2\% \\
  Red & 5314 & 15113 & 47.8\% & 20427 & 33.1\% \\  \hline
  Total & 13034 & 29698 & 100\% & 61712 & 100\% \\
  \hline
    \end{tabular}
  \end{center}
\end{table}


\subsubsection{Annotated Video Dataset (AVS)}

95 minutes of video has been recorded in the factory floor participating in this study (Figure~\ref{fig:example-stack-light-and-system}).
The video features five machines with
four different light models, shown in Figure~\ref{fig:datasets} (Top/Right).
Three models are tricoloured (green, yellow and red), and one has two colours (green and white).
The video is captured in QHD (2560$\times$1440) at 30 fps with a Logitech BRIO webcam. 
During the recording, the frame rate decreased slightly to an average of 22 fps due to the hardware heating up, resulting in 125613 frames and 628095 sub-images of stack lights.
The lights do not move, so the same bounding boxes (marked manually) can be used for every frame.
Stack lights are then cropped, adding portions of the background randomly from 0 to 20 pixels to each side.
This is done to introduce some context, and to increase diversity when training the classification network.
%
Each light is annotated with one of 9 labels according to the possible combinations of colours, including two or three colours lit simultaneously.
The labels and number of occurrences are listed in Table~\ref{tab:Occurrences-AVS-SLCD}.
As it can be observed, Green and GreenYellow are by far the most common classes.
%
%
Due to the high amount of frames to be labelled, the process was automated by training a stand-alone network
on a small number of images, and running it several times with manual control in each pass.
%

\begin{table}[htb]
  \begin{center}
    \caption{Occurrences of light combinations.}
    \label{tab:Occurrences-AVS-SLCD}
\begin{tabular}{ccccccc}


  \multicolumn{2}{c}{} & \multicolumn{2}{c}{AVS dataset} & \multicolumn{1}{c}{} & \multicolumn{2}{c}{SLCD dataset} \\ \cline{1-1} \cline{3-4} \cline{6-7}

  Combination &  & Occurrences & (\%) &  & Occurrences & (\%) \\  \cline{1-1} \cline{3-4} \cline{6-7}
  Green &  & 253253 & 40.3\% &  & 1060 & 9.6\% \\
  GreenRed &  & 29585 & 4.7\% &  & 1017 & 9.2\% \\
  GreenWhite &  & 17454 & 2.8\% &  & 1057 & 9.6\% \\
  GreenYellow &  & 198514 & 31.6\% &  & 1061 & 9.6\% \\
  GreenYellowRed &  & 10905 & 1.7\% &  & 916 & 8.3\% \\
  Yellow &  & 60124 & 9.6\% &  & 1114 & 10.1\% \\
  YellowRed &  & 11739 & 1.9\% &  & 1041 & 9.4\% \\
  Red &  & 21143 & 3.4\% &  & 1042 & 9.4\% \\
  off &  & 25348 & 4.0\% &  & 1067 & 9.7\% \\
  other objects &  & - & - &  & 1662 & 15.1\% \\ \cline{1-1} \cline{3-4} \cline{6-7}
  Total images &  & 628065 & 100\% &  & 11037 & 100\% \\  \cline{1-1} \cline{3-4} \cline{6-7}
  Total frames &  & 125613 & - &  & - & - \\  \cline{1-1} \cline{3-4} \cline{6-7}
\end{tabular}
  \end{center}
\end{table}

\subsubsection{Stack Light Classification Dataset (SLCD)}

This dataset has 11037 images extracted from the AVS dataset.
Every image has been double-checked manually to guarantee that is correctly labelled.
The images range in size from 25$\times$86 to 57$\times$123 pixels.
%
%
A 10$^{th}$ category of ``other objects'' has been added (see Table~\ref{tab:Occurrences-AVS-SLCD}), with images randomly cropped from the scene (Figure~\ref{fig:datasets}, Bottom/Left).
Its purpose is to enable detection of occlusions due to passing people or objects, or mitigate false positives given by the detector.
%
%

\begin{figure*}
\centering
\includegraphics[width=0.75\textwidth]{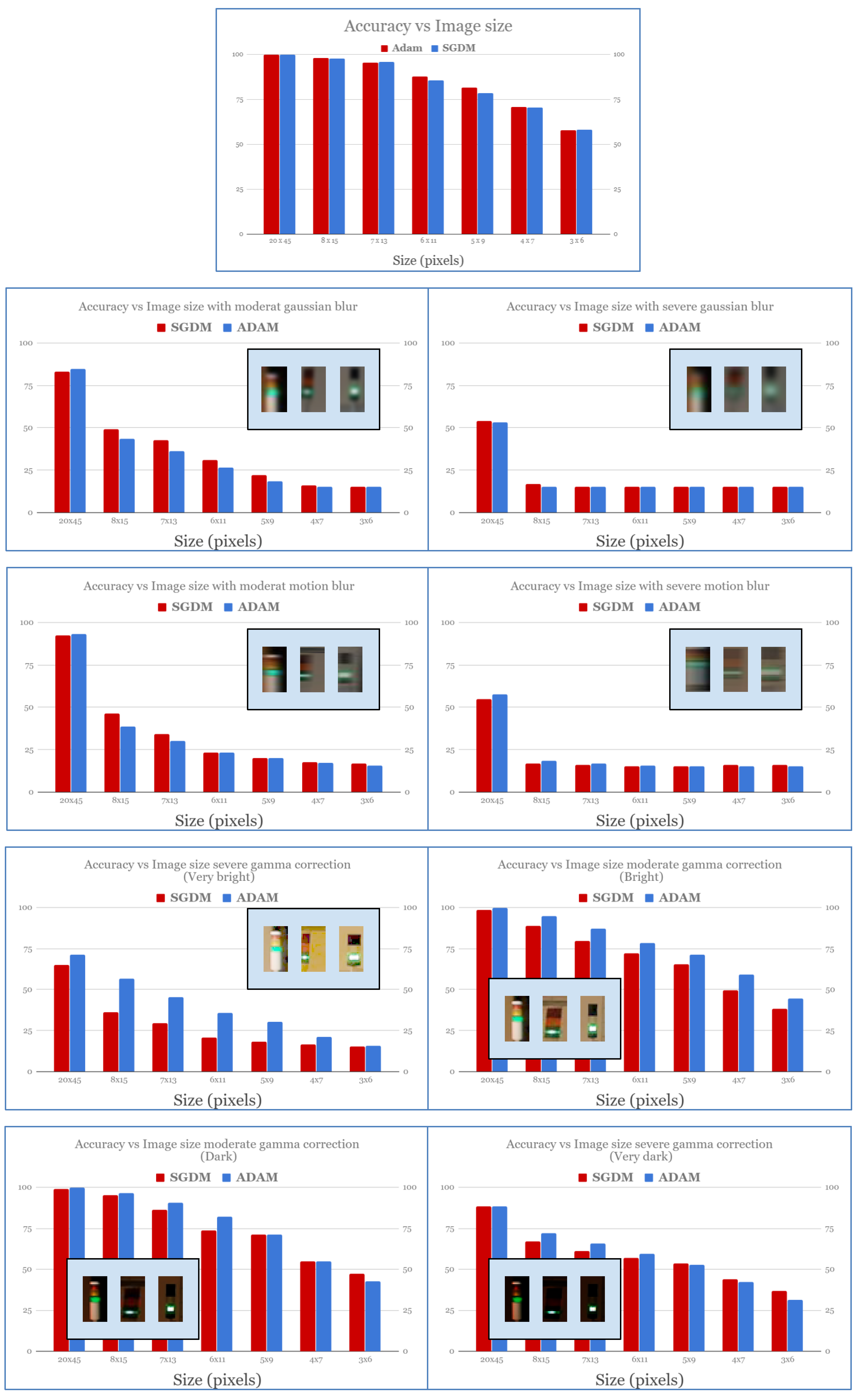}
\caption{Classification accuracy under different image perturbations.}
\label{fig:accuracy-classification}
\end{figure*}

\section{Experiments and Results}


\subsection{Detection}

%

The detection network YOLOv2 has been trained with 4840 frames from the AVS dataset.
As mentioned, frames are resized to 1248$\times$832 pixels, and split into six sub-images of 416$\times$416 pixels.
To increase variability in the training data, frames are cropped and resized several times, so that the lights appear at 0.75, 1, 2, and 4 times their actual size.
This is indicated by the yellow rectangles of Figure~\ref{fig:example-stack-light-and-system}, top.
The total number of sub-images of 416$\times$416 pixels after this process is 48400.
Differences occur mostly due to the colours that are lit in each frame, and people sitting/walking, so
after two epochs, the loss already decreased to $\sim$5$\times$10$^{-2}$, and the root-mean-square-error to $\sim$1.5$\times$10$^{-1}$, so training was stopped.
%
%
%
The detection network was then evaluated with 5120 unseen frames from the AVS dataset, containing 5$\times$5120=25600 stack lights.
%
%
%
There were 25387 true positives, 213 false negatives and 2 false positives. This means that 99.2\% of all lights are detected with a $<$0.1\% chance of a false detection.
The software was tested using an Nvidia GeForce 970 graphics card, with detection on an entire image taking about 0.40 seconds (2.5 fps).

\subsection{Classification}


The classification network AlexNet has been fine-tuned 
using the TLD dataset, with 80\%/20\% for training/validation.
The resulting network is then further fine-tuned using the SLCD dataset to recognise the ten classes of Table~\ref{tab:Occurrences-AVS-SLCD}.
The dataset has been split 60\%/20\%\/20\%\ for training/validation/testing. 
The network has been trained using both Adam and SGDM optimization, with similar results (99.4\% accuracy with Adam, 99.9\% with SGDM).
%
%
%
%
%
%
Some errors occur when lights are turning on or off (Figure~\ref{fig:datasets}, Bottom/Right),
so the network classifies the light as off when is on, or vice-versa.
%
%
The fading in/out process could be detected
by averaging the result across several frames,
or by thresholding the classification scores of the last layer of the network.
This is also due to the labelling employed, which only considers lights on or off,
but in reality they can have half/half state as well.
%
%
%
Misclassifications also occur with the ``other objects'' category containing bright lights.
For example, the last object shown in Figure~\ref{fig:datasets}, Bottom/Left, is erroneously classified as a green stack light.
It should be noted that the classification network is evaluated using labelled images from the SLCD dataset,
not the output given by the detection network. Thus, it could be expected that if the detection network has low false positives,
static objects from the ``other objects'' category will be discarded earlier.
%

We further analyze the capabilities of the classification network to operate under different image perturbations.
We use synthetically degraded images with the following perturbations: reduced image size, image blur caused
by defocus and by motion, and lightning variation.
%
%
To test different image sizes that simulate different distances to the camera,
images in the test set are resized with bicubic interpolation to
20$\times$45, 8$\times$15, 7$\times$13, 6$\times$11, 5$\times$9, 4$\times$7, and 3$\times$6.
Defocus blur is simulated
with a low-pass Gaussian filter of standard deviation 1.5 (moderate) and 3 (severe).
This simulates objects not correctly focused due to e.g. zooming or incorrect aperture.
Motion blur on the other hand results from the relative object/camera movement.
It is simulated 
with two parameters: direction (angle) and amount (strength).
The strength corresponds to the length of the blur in pixels.
For simplicity, we set the angle to zero
(blur along the horizontal axis), and strength to 10 pixels (moderate) and 20 pixels (severe).
Lastly, varying lightning conditions are modelled by applying gamma correction with a value of
%
%
$0.5$ and $0.25$ to obtain brighter images,
and $1.5$ and $2.5$ to obtain darker images.

Classification accuracy is given in Figure~\ref{fig:accuracy-classification} for the different types of degradations.
Regarding image size, accuracy starts to drop at lights sized 8$\times$15 pixels, and decreases faster after 6$\times$11 pixels. At a size of 3$\times$6 pixels, the accuracy has decreased to $\sim$57\%.
At this size, the colours start to be indistinguishable from each other.
All other degradations have been applied to the seven size groups.
With moderate defocus blur,
classification performance is reduced to $\sim$84\% for the group with biggest images (20$\times$45), and to $\sim$52\% in case of severe blur.
As images become smaller or increasingly blurry, the accuracy decreases further up to $\sim$15\%. With severe blur, accuracy is already significantly low for
the second image size group (8$\times$15).
A similar behaviour is observed in presence of motion blur.
Therefore, the classification network is very sensitive to these two types of blur,
which is further accentuated with smaller image sizes.
On the other hand, the impact of lightning variation is not so severe, at least under
moderate levels of gamma correction.
With moderate levels, accuracy remains close to 99\% for the group with biggest images (20$\times$45),
both if the image is darkened or brightened.
Down to 7$\times$13 pixels, accuracy still remains above 90\%, and above 75\% for images of 6$\times$11 pixels.
With severe gamma correction, on the other hand, accuracy decreases very quickly, even for the group with biggest images.


%

\section{Conclusion}

This paper has presented a system that makes use of computer vision to detect and classify lights in industrial signal lights towers, also called stack lights.
These indicate the operational state of each machine. Measuring them automatically can provide important benefits to streamline production processes in factories,
e.g. detect maintenance or performance issues, or help to reconfigure the production efficiently.
We have split our solution into one classification and one detection network.
We have trained and evaluated YOLOv2 \cite{[Redmon17YOLOv2]} as detection network, and AlexNet \cite{[Krizhevsky12]} as classification network.
A dataset with 125613 frames and 628095 sub-images of stack lights has been also captured
in the industrial factory floor participating in this study (Figure~\ref{fig:example-stack-light-and-system}).
The presented solution performs well in the pre-described environment (indoor and no motion between camera and lights),
with both a detection and classification accuracy of $\sim$99\%.
%

The classification network employed is less heavy to run in comparison to the detection network,
hence the split in two steps.
YOLO could theoretically be trained to perform classifications as well, but that would require a more powerful hardware (GPU) \cite{[Redmon16YOLO],[Redmon17YOLOv2]}.
The same applies to architectures based on R-CNN \cite{[Girshick14RCNN]}.
Our approach allowed the proposed system to be run on a regular laptop, and would be transferrable to setups with limited computing resources.
The classification network is also evaluated under different image perturbations simulating small image size, defocus blur, motion blur, and lightning changes.
These try to simulate other operational conditions different than the ones found in our factory setup, such as outdoor, distant acquisition,
or motion.
Machines are becoming increasingly mobile and for some industries, production outdoors can occur.
It has been observed that
scale changes do not produce a significant reduction in accuracy up to a size of 7$\times$13 pixels,
where lights can be classified reliably with an accuracy of $\sim$97\%.
After this, accuracy decreases rapidly and at a resolution of 3$\times$6, accuracy is $\sim$58\%.
Defocus and gaussian blur have a more significant effect.
Moderate levels of blur already bring accuracy down to $\sim$84\%, even if image size is kept high (20$\times$45).
As images become smaller or increasingly blurry, the accuracy decreases significantly more.
Moderate lightning changes, on the other hand, allow to reduce image size to a certain extent
without sacrificing accuracy significantly.
Performing retraining on data with the perturbations employed here could get the network more
accurate \cite{[Pan10]}, and will be the source of future work.
Also, although a large number of frames could be collected, all of them picture the same scene from the same environment.
This was due to privacy and operational restrictions at the factory when the database was captured.
In this setup we have shown that classification of factory signal lights can be automated with high accuracy.
Future areas of research include studies with a richer dataset from different locations, view angles, containing motion, etc.
that could be used to produce a solution that generalizes to more diverse environments.
%
%

\section*{Acknowledgment}

Author F. A.-F. thanks the Swedish Research Council and
the CAISR program of the Swedish Knowledge Foundation for funding his research.
Authors also acknowledge the support of HMS Industrial Networks AB.

\end{document}